# Towards a single proposal in spelling correction

Eneko Agirre, Koldo Gojenola, Kepa Sarasola
Dept. of Computer Languages and Systems
University of the Basque Country, 649 P. K.,
E-20080 Donostia, Basque Country
eneko@si.ehu.es

Atro Voutilainen
Department of General Linguistics
University of Helsinki, P.O. Box 4
FIN-00014 Helsinki, Finland
avoutila@ling.helsinki.fi

## Abstract

The study presented here relies on the integrated use of different kinds of knowledge in order to improve first-guess accuracy in non-word context-sensitive correction for general unrestricted texts. State of the art spelling correction systems, e.g. *ispell*, apart from detecting spelling errors, also assist the user by offering a set of candidate corrections that are close to the misspelled word. Based on the correction proposals of *ispell*, we built several guessers, which were combined in different ways. Firstly, we evaluated all possibilities and selected the best ones in a corpus with artificially generated typing errors. Secondly, the best combinations were tested on texts with genuine spelling errors. The results for the latter suggest that we can expect automatic non-word correction for *all* the errors in a free running text with 80% precision and a single proposal 98% of the times (1.02 proposals on average).

## Introduction

The problem of devising algorithms and techniques for automatically correcting words in text remains a research challenge. Existing spelling correction techniques are limited in their scope and accuracy. Apart from detecting spelling errors, many programs assist users by offering a set of candidate corrections that are close to the misspelled word. This is true for most commercial word-processors as well as the Unix-based spelling-corrector *ispell*[1] (1993). These programs tolerate lower first guess accuracy by returning multiple guesses, allowing the user to make the final choice of the intended word. In contrast, some applications will require fully automatic correction for general-purpose texts (Kukich 1992).

It is clear that context-sensitive spelling correction offers better results than isolated-word error correction. The underlying task is to determine the relative degree of well formedness among alternative sentences (Mays et al. 1991). The question is what kind of knowledge (lexical, syntactic, semantic, ...) should be represented, utilised and combined to aid in this determination.

This study relies on the integrated use of three kinds of knowledge (syntagmatic, paradigmatic and statistical) in order to improve first guess accuracy in non-word context-sensitive correction for general unrestricted texts. Our techniques were applied to the corrections posed by *ispell*. Constraint Grammar (Karlsson et al. 1995) was chosen to represent syntagmatic knowledge. Its use as a part of speech tagger for English has been highly successful. Conceptual Density (Agirre and Rigau 1996) is the paradigmatic component chosen to discriminate semantically among potential noun corrections. This technique measures "affinity distance" between nouns using Wordnet (Miller 1990). Finally, general and document word-occurrence frequency-rates complete the set of knowledge sources combined. We knowingly did not use any model of common misspellings, the main reason being that we did not want to use knowledge about the error source. This work focuses on language models, not error models (typing errors, common misspellings, OCR mistakes, speech recognition mistakes, etc.).

The system was evaluated against two sets of texts: artificially generated errors from the Brown corpus (Francis and Kucera 1967) and genuine spelling errors from the Bank of English[2].

The remainder of this paper is organised as follows. Firstly, we present the techniques that will be evaluated and the way to combine them. Section 2 describes the experiments and shows the results, which are evaluated in section 3. Section 4 compares other relevant work in context sensitive correction.

---

[1] *Ispell* was used for the spell-checking and correction candidate generation. Its assets include broad-coverage and excellent reliability.

[2] http://titania.cobuild.collins.co.uk/boe_info.html

# 1    The basic techniques

## 1.1   Constraint Grammar (CG)

Constraint Grammar was designed with the aim of being a language-independent and robust tool to disambiguate and analyse unrestricted texts. CG grammar statements are close to real text sentences and directly address parsing problems such as ambiguity. Its application to English (ENGCG[3]) resulted a very successful part of speech tagger for English. CG works on a text where all possible morphological interpretations have been assigned to each word-form by the ENGTWOL morphological analyser (Voutilainen and Heikkilä 1995). The role of CG is to apply a set of linguistic constraints that discard as many alternatives as possible, leaving at the end almost fully disambiguated sentences, with one morphological or syntactic interpretation for each word-form. The fact that CG tries to leave a unique interpretation for each word-form makes the formalism adequate to achieve our objective.

### *Application of Constraint Grammar*

The text data was input to the morphological analyser. For each unrecognised word, *ispell* was applied, placing the morphological analyses of the correction proposals as alternative interpretations of the erroneous word (see example 1). EngCG-2 morphological disambiguation was applied to the resulting texts, ruling out the correction proposals with an incompatible POS (cf. example 2). We must note that the broad coverage lexicons of *ispell* and ENGTWOL are independent. This caused the correspondence between unknown words and *ispell*'s proposals not to be one to one with those of the EngCG-2 morphological analyser, especially in compound words. Such problems were solved considering that a word was correct if it was covered by any of the lexicons.

## 1.2   Conceptual Density (CD)

The discrimination of the correct category is unable to distinguish among readings belonging to the same category, so we also applied a word-sense disambiguator based on Wordnet, that had already been tried for nouns on free-running text. In our case it would choose the correction proposal semantically closer to the surrounding context. It has to be noticed that Conceptual Density can only be applied when all the proposals are categorised as nouns, due to the structure of Wordnet.

```
<our>
   "our" PRON PL ...
<bos> ; INCORRECT OR SPELLING ERROR
   "boss" N S
   "boys" N P
   "bop" V S
   "Bose" <Proper>
```

**Example 1. Proposals and morphological analysis
for the misspelling *bos***

```
<our>
   "our" PRON PL ...
<bos> ; INCORRECT OR SPELLING ERROR
   "boss" N S
   "boys" N P
   "bop" V S
   "Bose" <Proper>
<are>            ...
```

**Example 2. CG leaves only nominal proposals**

## 1.3   Frequency statistics (DF & BF)

Frequency data was calculated as word-form frequencies obtained from the document where the error was obtained (Document frequency, DF) or from the rest of the documents in the whole Brown Corpus (Brown frequency, BF). The experiments proved that word-forms were better suited for the task, compared to frequencies on lemmas.

## 1.4   Other interesting heuristics (H1, H2)

We eliminated proposals beginning with an uppercase character when the erroneous word did not begin with uppercase and there were alternative proposals beginning with lowercase. In example 1, the fourth reading for the misspelling "bos" was eliminated, as "Bose" would be at an editing distance of two from the misspelling (heuristic H1). This heuristic proved very reliable, and it was used in all experiments. After obtaining the first results, we also noticed that words with less than 4 characters like "si", "teh", ... (misspellings for "is" and "the") produced too many proposals, difficult to disambiguate. As they were one of the main error sources for our method, we also evaluated the results excluding them (heuristic H2).

---

## 1.5 Combination of the basic techniques using votes

We considered all the possible combinations among the different techniques, e.g. CG+BF, BF+DF, and CG+DF. The weight of the vote can be varied for each technique, e.g. CG could have a weight of 2 and BF a weight of 1 (we will represent this combination as CG2+BF1). This would mean that the BF candidate(s) will only be chosen if CG does not select another option or if CG selects more than one proposal. Several combinations of weights were tried. This simple method to combine the techniques can be improved using optimization algorithms to choose the best weights among fractional values. Nevertheless, we did some trials weighting each technique with its expected precision, and no improvement was observed. As the best combination of techniques and weights for a given set of texts can vary, we separated the error corpora in two, trying all the possibilities on the first half, and testing the best ones on the second half (c.f. section 2.1).

## 2 The experiments

Based on each kind of knowledge, we built simple guessers and combined them in different ways. In the first phase, we evaluated all the possibilities and selected the best ones on part of the corpus with artificially generated errors. Finally, the best combinations were tested against the texts with genuine spelling errors.

### 2.1 The error corpora

We chose two different corpora for the experiment. The first one was obtained by systematically generating misspellings from a sample of the Brown Corpus, and the second one was a raw text with genuine errors. While the first one was ideal for experimenting, allowing for automatic verification, the second one offered a realistic setting. As we said before, we are testing language models, so that both kinds of data are appropriate. The corpora with artificial errors, artificial corpora for short, have the following features: a sample was extracted from SemCor (a subset of the Brown Corpus) selecting 150 paragraphs at random. This yielded a seed corpus of 505 sentences and 12659 tokens. To simulate spelling errors, a program named *antispell,* which applies Damerau's rules at random, was run, giving an average of one spelling error for each 20 words (non-words were left untouched). *Antispell* was run 8 times on the seed corpus, creating 8 different corpora with the same text but different errors. Nothing was done to prevent two errors in the same sentence, and some paragraphs did not have any error.

The corpus of genuine spelling errors, which we also call the "real" corpus for short, was magazine text from the Bank of English Corpus, which probably was not previously spell-checked (it contained many misspellings), so it was a good source of errors. Added to the difficulty of obtaining texts with real misspellings, there is the problem of marking the text and selecting the correct proposal for automatic evaluation.

As mentioned above, the artificial-error corpora were divided in two subsets. The first one was used for training purposes[4]. Both the second half and the "real" texts were used for testing.

### 2.2 Data for each corpora

The two corpora were passed trough *ispell*, and for each unknown word, all its correction proposals were inserted. Table 1 shows how, if the misspellings are generated at random, 23.5% of them are real words, and fall out of the scope of this work. Although we did not make a similar counting in the real texts, we observed that a similar percentage can be expected.

| | 1st half | 2nd half | "real" |
|---|---|---|---|
| words | 47584 | 47584 | 39733 |
| errors | 1772 | 1811 | _[5] |
| non real-word errors | 1354 | 1403 | 369 |
| ispell proposals | 7242 | 8083 | 1257 |
| words with multiple proposals | 810 | 852 | 158 |
| long word errors (H2) | 968 | 980 | 331 |
| proposals for long words (H2) | 2245 | 2313 | 807 |
| long word errors (H2) with multiple proposals | 430 | 425 | 124 |

**Table 1. Number of errors and proposals**

For the texts with genuine errors, the method used in the selection of the misspellings was the following: after applying *ispell*, no correction was found for 150 words (mainly proper nouns and foreign words), and there were about 300 which were formed by joining two consecutive words or by special affixation rules (*ispell* recognised them

---

[4] In fact, there is no training in the statistical sense. It just involves choosing the best alternatives for voting.

[5] As we focused on non-word words, there is not a count of real-word errors.

| | Cover.% | Prec.% | #prop. |
|---|---|---|---|
| **Basic techniques** | | | |
| random baseline | 100.00 | 54.36 | 1.00 |
| random+H2 | 71.49 | 71.59 | 1.00 |
| CG | 99.85 | 86.91 | 2.33 |
| CG+H2 | 71.42 | 95.86 | 1.70 |
| BF | 96.23 | 86.57 | 1.00 |
| BF+H2 | 68.69 | 92.15 | 1.00 |
| DF | 90.55 | 89.97 | 1.02 |
| DF+H2 | 62.92 | 96.13 | 1.01 |
| CD | 6.06 | 79.27 | 1.01 |
| **Combinations** | | | |
| CG1+DF2 | 99.93 | 90.39 | 1.17 |
| CG1+DF2+H2 | 71.49 | 96.38 | 1.12 |
| CG1+DF1+BF1 | 99.93 | 89.14 | 1.03 |
| CG1+DF1+BF1+H2 | 71.49 | 94.73 | 1.03 |
| CG1+DF1+BF1+CD1 | 99.93 | 89.14 | 1.02 |
| CG1+DF1+BF1+CD1+H2 | 71.49 | 94.63 | 1.02 |

**Table 2. Results for several combinations (1ˢᵗ half)**

| | Cover. | Prec. | #prop |
|---|---|---|---|
| **Basic techniques** | | | |
| random baseline | 100.00 | 23.70 | 1.00 |
| random+H2 | 52.70 | 36.05 | 1.00 |
| CG | 99.75 | 78.09 | 3.23 |
| CG+H2 | 52.57 | 90.68 | 2.58 |
| BF | 93.70 | 76.94 | 1.00 |
| BF+H2 | 48.04 | 81.38 | 1.00 |
| DF | 84.20 | 81.96 | 1.03 |
| DF+H2 | 38.48 | 89.49 | 1.03 |
| CD | 8.27 | 75.28 | 1.01 |
| **Combinations** | | | |
| CG1+DF2 | 99.88 | 83.93 | 1.28 |
| CG1+DF2+H2 | 52.70 | 91.86 | 1.43 |
| CG1+DF1+BF1 | 99.88 | 81.83 | 1.04 |
| CG1+DF1+BF1+H2 | 52.70 | 88.14 | 1.06 |
| CG1+DF1+BF1+CD1 | 99.88 | 81.83 | 1.04 |
| CG1+DF1+BF1+CD+H2 | 52.70 | 87.91 | 1.05 |

**Table 3. Results on errors with multiple proposals (1ˢᵗ half)**

| | Cover.% | Prec.% | #prop |
|---|---|---|---|
| **Basic techniques** | | | |
| random baseline | 100.00 | 53.67 | 1.00 |
| random+H2 | 69.85 | 71.53 | 1.00 |
| DF | 90.31 | 89.50 | 1.02 |
| DF+H2 | 61.51 | 95.60 | 1.01 |
| **Combinations** | | | |
| CG1+DF2 | 99.64 | 90.06 | 1.19 |
| CG1+DF2+H2 | 69.85 | 95.71 | 1.22 |
| CG1+DF1+BF1 | 99.64 | 87.77 | 1.03 |
| CG1+DF1+BF1+H2 | 69.85 | 93.16 | 1.03 |
| CG1+DF1+BF1+CD1 | 99.64 | 87.91 | 1.03 |
| CG1+DF1+BF1+CD+H2 | 69.85 | 93.27 | 1.02 |

**Table 4. Validation of the best combinations (2ⁿᵈ half)**

| | Cover. | Prec. | #pro |
|---|---|---|---|
| **Basic techniques** | | | |
| random baseline | 100.00 | 23.71 | 1.00 |
| random+H2 | 50.12 | 34.35 | 1.00 |
| DF | 84.04 | 81.42 | 1.03 |
| DF+H2 | 36.32 | 87.66 | 1.04 |
| **Combinations** | | | |
| CG1+DF2 | 99.41 | 83.59 | 1.31 |
| CG1+DF2+H2 | 50.12 | 90.12 | 1.50 |
| CG1+DF1+BF1 | 99.41 | 79.81 | 1.05 |
| CG1+DF1+BF1+H2 | 50.12 | 84.24 | 1.06 |
| CG1+DF1+BF1+CD1 | 99.41 | 80.05 | 1.05 |
| CG1+DF1+BF1+CD1+H2 | 50.12 | 84.47 | 1.06 |

**Table 5. Results on errors with multiple proposals (2ⁿᵈ half)**

## 2.3 Results

We mainly considered three measures:

- coverage: the number of errors for which the technique yields an answer.
- precision: the number of errors with the correct proposal among the selected ones
- remaining proposals: the average number of selected proposals.

### 2.3.1 Search for the best combinations

Table 2 shows the results on the training corpora. We omit many combinations that we tried, for the sake of brevity. As a baseline, we show the results when the selection is done at random. Heuristic H1 is applied in all the cases, while tests are performed with and without heuristic H2. If we focus on the errors for which *ispell* generates more than one correction proposal (cf. table 3), we get a better estimate of the contribution of each guesser. There were 8.26 proposals per word in the general correctly). This left 369 erroneous word-forms. After examining them we found that the correct word-form was among *ispell*'s proposals, with very few exceptions. Regarding the selection among the different alternatives for an erroneous word-form, we can see that around half of them has a single proposal. This gives a measure of the work to be done. For example, in the real error corpora, there were 158 word-forms with 1046 different proposals. This means an average of 6.62 proposals per word. If words of length less than 4 are not taken into account, there are 807 proposals, that is, 4.84 alternatives per word.

| | Cover. % | Prec. % | #prop. |
|---|---|---|---|
| **Basic techniques** | | | |
| random baseline | 100.00 | 69.92 | 1.00 |
| random+H2 | 89.70 | 75.47 | 1.00 |
| CG | 99.19 | 84.15 | 1.61 |
| CG+H2 | 89.43 | 90.30 | 1.57 |
| DF | 70.19 | 93.05 | 1.02 |
| DF+H2 | 61.52 | 97.80 | 1.00 |
| BF | 98.37 | 80.99 | 1.00 |
| BF+H2 | 88.08 | 85.54 | 1.00 |
| **Combinations** | | | |
| CG1+DF2 | 100.00 | 87.26 | 1.42 |
| CG1+DF2+H2 | 89.70 | 90.94 | 1.43 |
| CG1+DF1+BF1 | 100.00 | 80.76 | 1.02 |
| CG1+DF1+BF1+H2 | 89.70 | 84.89 | 1.02 |

**Table 6. Best combinations ("real" corpus)**

| | Cover. % | Prec. % | #prop |
|---|---|---|---|
| **Basic techniques** | | | |
| random baseline | 100.00 | 29.75 | 1.00 |
| random+H2 | 76.54 | 34.52 | 1.00 |
| CG | 98.10 | 62.58 | 2.45 |
| CG+H2 | 75.93 | 73.98 | 2.52 |
| DF | 30.38 | 62.50 | 1.13 |
| DF+H2 | 12.35 | 75.00 | 1.05 |
| BF | 96.20 | 54.61 | 1.00 |
| BF+H2 | 72.84 | 60.17 | 1.00 |
| **Combinations** | | | |
| CG1+DF2 | 100.00 | 70.25 | 1.99 |
| CG1+DF2+H2 | 76.24 | 75.81 | 2.15 |
| CG1+DF1+BF1 | 100.00 | 55.06 | 1.04 |
| CG1+DF1+BF1+H2 | 76.54 | 59.68 | 1.05 |

**Table 7. Results on errors with multiple proposals ("real" corpus)**

case, and 3.96 when H2 is applied. The results for all the techniques are well above the random baseline. The single best techniques are DF and CG. CG shows good results on precision, but fails to choose a single proposal. H2 raises the precision of all techniques at the cost of losing coverage. CD is the weakest of all techniques, and we did not test it with the other corpora. Regarding the combinations, CG1+DF2+H2 gets the best precision overall, but it only gets 52% coverage, with 1.43 remaining proposals. Nearly 100% coverage is attained by the H2 combinations, with highest precision for CG1+DF2 (83% precision, 1.28 proposals).

### 2.3.2 Validation of the best combinations

In the second phase, we evaluated the best combinations on another corpus with artificial errors. Tables 4 and 5 show the results, which agree with those obtained in 2.3.1. They show slightly lower percentages but always in parallel.

### 2.3.3 Corpus of genuine errors

As a final step we evaluated the best combinations on the corpus with genuine typing errors. Table 6 shows the overall results obtained, and table 7 the results for errors with multiple proposals. For the latter there were 6.62 proposals per word in the general case (2 less than in the artificial corpus), and 4.84 when heuristic H2 is applied (one more that in the artificial corpus). These tables are further commented in the following section.

## 3 Evaluation of results

This section reviews the results obtained. The results for the "real" corpus are evaluated first, and the comparison with the other corpora comes later. Concerning the application of each of the simple techniques separately[6]:

- Any of the guessers performs much better than random.

- DF has a high precision (75%) at the cost of a low coverage (12%). The difference in coverage compared to the artificial error corpora (84%) is mainly due to the smaller size of the documents in the real error corpus (around 50 words per document). For medium-sized documents we expect a coverage similar to that of the artificial error corpora.

- BF offers lower precision (54%) with the gains of a broad coverage (96%).

- CG presents 62% precision with nearly 100% coverage, but at the cost of leaving many proposals (2.45)

- The use of CD works only with a small fraction of the errors giving modest results. The fact that it was only applied a few times prevents us from making further conclusions.

Combining the techniques, the results improve:

- The CG1+DF2 combination offers the best results in coverage (100%) and precision (70%) for all tests. As can be seen, CG raises the coverage of the DF method, at the cost of also increasing the number of proposals (1.9) per erroneous word. Had the coverage of DF increased, so would also the number of

---

proposals decrease for this combination, for instance, close to that of the artificial error corpora (1.28).

- The CG1+DF1+BF1 combination provides the same coverage with nearly one interpretation per word, but decreasing precision to a 55%.

- If full coverage is not necessary, the use of the H2 heuristic raises the precision at least 4% for all combinations.

When comparing these results with those of the artificial errors, the precisions in tables 2, 4 and 6 can be misleading. The reason is that the coverage of some techniques varies and the precision varies accordingly. For instance, coverage of DF is around 70% for real errors and 90% for artificial errors, while precisions are 93% and 89% respectively (cf. tables 6 and 2). This increase in precision is not due to the better performance of DF[7], but can be explained because the lower the coverage, the higher the proportion of errors with a single proposal, and therefore the higher the precision.

The comparison between tables 3 and 7 is more clarifying. The performance of all techniques drops in table 7. Precision of CG and BF drops 15 and 20 points. DF goes down 20 points in precision and 50 points in coverage. This latter degradation is not surprising, as the length of the documents in this corpus is only of 50 words on average. Had we had access to medium sized documents, we would expect a coverage similar to that of the artificial error corpora.

The best combinations hold for the "real" texts, as before. The highest precision is for CG1+DF2 (with and without H2). The number of proposals left is higher in the "real" texts than in the artificial ones (1.99 to 1.28). It can be explained because DF does not manage to cover all errors, and that leaves many CG proposals untouched.

We think that the drop in performance for the "real" texts was caused by different factors. First of all, we already mentioned that the size of the documents strongly affected DF. Secondly, the nature of the errors changes: the algorithm to produce spelling errors was biased in favour of frequent words, mostly short ones. We will have to analyse this question further, specially regarding the origin of the natural errors. Lastly,

---

[7] In fact the contrary is deduced from tables 3 and 7.

BF was trained on the Brown corpus on American English, while the "real" texts come from the Bank of English. Presumably, this could have also affected negatively the performance of these algorithms.

Back to table 6, the figures reveal which would be the output of the correction system. Either we get a single proposal 98% of the times (1.02 proposals left on average) with 80% precision for all non-word errors in the text (CG1+DF1+BF1) or we can get a higher precision of 90% with 89% coverage and an average of 1.43 proposals (CG1+DF2+H2).

# 4 Comparison with other context-sensitive correction systems

There is not much literature about automatic spelling correction with a single proposal. Menezo et al. (1996) present a spelling/grammar checker that adjusts its strategy dynamically taking into account different lexical agents (dictionaries, ...), the user and the kind of text. Although no quantitative results are given, this is in accord with using document and general frequencies.

Mays et al. (1991) present the initial success of applying word trigram conditional probabilities to the problem of context based detection and correction of real-word errors.

Yarowsky (1994) experiments with the use of decision lists for lexical ambiguity resolution, using context features like local syntactic patterns and collocational information, so that multiple types of evidence are considered in the context of an ambiguous word. In addition to word-forms, the patterns involve POS tags and lemmas. The algorithm is evaluated in missing accent restoration task for Spanish and French text, against a predefined set of a few words giving an accuracy over 99%.

Golding and Schabes (1996) propose a hybrid method that combines part-of-speech trigrams and context features in order to detect and correct real-word errors. They present an experiment where their system has substantially higher performance than the grammar checker in MS Word, but its coverage is limited to eighteen particular confusion sets composed by two or three similar words (e.g.: weather, whether).

The last three systems rely on a previously collected set of confusion sets (sets of similar words or accentuation ambiguities). On the contrary, our system has to choose a single proposal for any possible spelling error, and it is therefore impossible to collect the confusion sets (i.e. sets of proposals for each spelling error) beforehand. We also need to correct as many errors as possible, even if the amount of data for a particular case is scarce.

## Conclusion

This work presents a study of different methods that build on the correction proposals of *ispell*, aiming at giving a single correction proposal for misspellings. One of the difficult aspects of the problem is that of testing the results. For that reason, we used both a corpus with artificially generated errors for training and testing, and a corpus with genuine errors for testing.

Examining the results, we observe that the results improve as more context is taken into account. The word-form frequencies serve as a crude but helpful criterion for choosing the correct proposal. The precision increases as closer contexts, like document frequencies and Constraint Grammar are incorporated. From the results on the corpus of genuine errors we can conclude the following. Firstly, the correct word is among *ispell*'s proposals 100% of the times, which means that all errors can be recovered. Secondly, the expected output from our present system is that it will correct automatically the spelling errors with either 80% precision with full coverage or 90% precision with 89% coverage and leaving an average of 1.43 proposals.

Two of the techniques proposed, Brown Frequencies and Conceptual Density, did not yield useful results. CD only works for a very small fraction of the errors, which prevents us from making further conclusions.

There are reasons to expect better results in the future. First of all, the corpus with genuine errors contained very short documents, which caused the performance of DF to degrade substantially. Further tests with longer documents should yield better results. Secondly, we collected frequencies from an American English corpus to correct British English texts. Once this language mismatch is solved, better performance should be obtained. Lastly, there is room for improvement in the techniques themselves. We knowingly did not use any model of common misspellings. Although we expect limited improvement, stronger methods to combine the techniques can also be tried.

Continuing with our goal of attaining a single proposal as reliably as possible, we will focus on short words and we plan to also include more syntactic and semantic context in the process by means of collocational information. This step opens different questions about the size of the corpora needed for accessing the data and the space needed to store the information.

## Acknowledgements

This research was supported by the Basque Government, the University of the Basque Country and the CICYT (Comisión Interministerial de Ciencia y Tecnología).